\documentclass{proposalnsf}
\usepackage{times}
\usepackage{epsfig}
\usepackage{graphicx}
\usepackage{amsfonts}       
\usepackage{amsmath}
\usepackage{amssymb}
\usepackage{amsthm}
\usepackage{commath}
\usepackage{bm}
\usepackage{url}
\newcommand{\comment}[1]{}

\title{\Large \bf NeRFs: The Search for the Best 3D Representation}
\author{Ravi Ramamoorthi \\ University of California San Diego\footnote{Email: ravir@cs.ucsd.edu.  The author is also currently affiliated wtih NVIDIA.}}
\date{}

\begin{document}
\maketitle

\vspace*{-0.2in}
\section*{Abstract}
Neural Radiance Fields or NeRFs have become the representation of choice for problems in view synthesis or image-based rendering, as well as in many other applications across computer graphics and vision, and beyond.  At their core, NeRFs describe a new representation of 3D scenes or 3D geometry.  Instead of meshes, disparity maps, multiplane images or even voxel grids, they represent the scene as a continuous volume, with volumetric parameters like view-dependent radiance and volume density obtained by querying a neural network.  The NeRF representation has now been widely used, with thousands of papers extending or building on it every year, multiple authors and websites providing overviews and surveys, and numerous industrial applications and startup companies.  In this article, we briefly review the NeRF representation, and describe the three decades-long quest to find the best 3D representation for view synthesis and related problems, culminating in the NeRF papers.  
We then describe new developments in terms of NeRF representations and make some observations and insights regarding the future of 3D representations.

\section{Introduction and Basic NeRF Algorithm}
This article is written in response to the Frontiers of Science Award generously granted in 2023 to the papers on ``NeRF: Representing Scenes as Neural Radiance Fields for View Synthesis''~\cite{nerf,nerfcacm} at ECCV 2020 and CACM 2022.  NeRFs were introduced as a method that achieved the highest quality visual and quantitative results to date for the task of view synthesis or image-based rendering.  This problem is easy to state, although notoriously difficult to solve well.  Given a number of input images of a 3D scene of interest, we seek to optimize some kind of internal representation, which will then allow us to synthesize novel views.  Achieving this enables one to take a few hand-held images of an object or scene and be able to sense it in 3D instead of as a flat 2D image.  There are numerous applications in immersive virtual and augmented reality.  As such, view synthesis enables promoting our standard 2D images or photographs to full 3D representations, also enabling newer applications like digital twins and the metaverse.  For example, NeRFs have already been used~\cite{LumaAI,GoogleMaps} in Google's Maps and StreetView to create immersive renderings from source photographs of cities, buildings and streets, and within the Luma AI mobile phone app to take a few images of an object of interest and produce a compelling fly-through.\footnote{The student co-first authors of the original papers now work at Google (Srinivasan, Mildenhall) and Luma AI (Tancik), as of this writing on July 2023.  The author of this article is not affiliated with either company.}  

\begin{figure}
\includegraphics[width=\columnwidth]{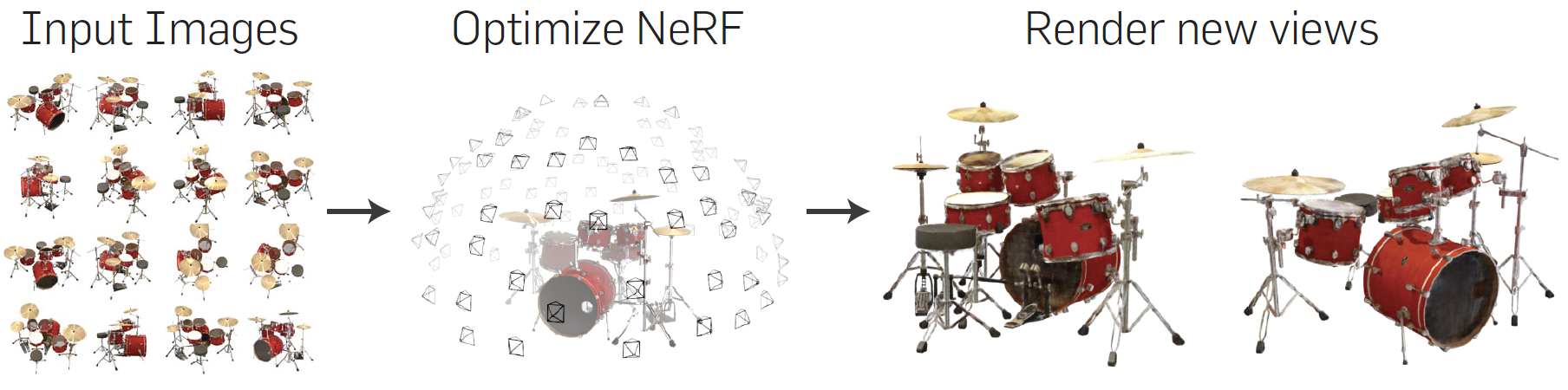}
\caption{\em The canonical view synthesis problem.  Given a set of input images (here 100 views of the synthetic drums scene randomly captured on the surrounding hemisphere), we first optimize a NeRF representation, then use it with volume rendering to synthesize two novel views, including view-dependent specular effects and reflections.  For the past three decades, a quest in view synthesis or image-based rendering has been to find the ``best'' intermediate representation for 3D scenes that will enable the highest-quality view synthesis; this figure could generically represent many previous papers in the area.}
\label{fig:setup}
\end{figure}

\begin{figure}
\includegraphics[width=\columnwidth]{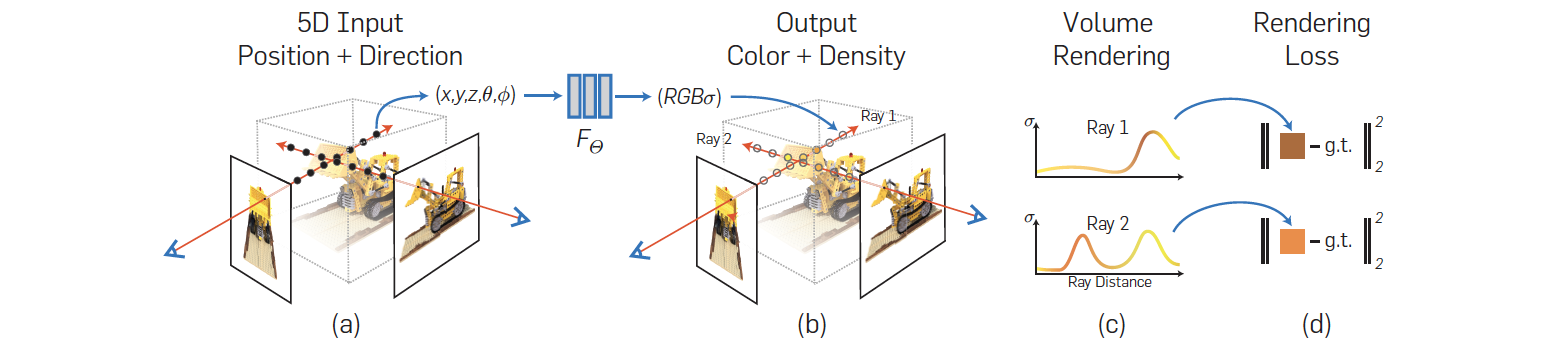}
\caption{\em Optimizing our neural radiance field representation is done simply by computing a simple loss or error between the reconstructed image computed using volume rendering and held-out views.  Medium parameters are determined by a multi-layer perceptron (MLP) that takes 5D coordinates as input and outputs view-dependent emissive radiance (rgb) and volume density ($\sigma$). }
\label{fig:optimization}
\end{figure}

NeRFs represent a 3D scene using a fully-connected neural network, known as a multi-layer perceptron or MLP.  As opposed to the commonly used convolutional neural network or CNN that has made possible many advances in artificial intelligence (AI) in recent years~\cite{Krizhevsky}, this representation is non-convolutional and in recent work is often not even a particularly-deep network, enabling simpler and more efficient algorithms that can be more easily trained and better suit the problem at hand.  In particular, the input is a single continuous 5D coordinate, involving the spatial position $(x,y,z)$ and the viewing direction $(\theta,\phi)$ and the output is the volume density $\sigma$ (essentially the absorption or extinction coefficient in the volume, since there is no scattering) and the view-dependent emitted radiance or loosely color $\bm c$.  Views are synthesized by querying 5D coordinates along camera rays and using standard volume rendering techniques well known in computer graphics and other fields to project these output colors and densities to an image.  A key aspect of deep learning or optimization is to be able to differentiate the relevant functions to run a gradient-descent optimizer.  Volume rendering is naturally differentiable (no discontinuities), so the only input needed is the set of images with known camera poses (usually obtained from a system such as COLMAP~\cite{colmap}).  Optimization is done simply by comparing the reconstructed value to the ``ground truth'' input images.  In many respects this is a significantly simpler pipeline than previous methods.  We show the setup of the problem and the optimization performed in Figs.~\ref{fig:setup} and~\ref{fig:optimization} (figures taken from~\cite{nerfcacm}).  

More formally, the color $C(\bm r)$ of camera ray $\bm r(t) = \bm o + t \bm d$ with origin (camera) $\bm o$ and direction $\bm d$ ($t$ is the distance parameter) with near and far bounds $t_n$ and $t_f$ in the volume is given by~\cite{nerfcacm},
\begin{equation}
C(\bm r) = \int_{t_n}^{t_f} T(t) \sigma(\bm r(t)) \bm c (\bm r(t), \bm d)\, dt,
\end{equation}
where $T(t)$ is the attenuation of the emissive radiance $\bm c$ at $t$ from traveling through the medium, $\sigma$ is the volume density and $\bm c$ is the view-dependent emissive color.\footnote{It is possible to bake/pre-multiply $\sigma$ into the color $\bm c$ but this does not provide as elegant a formulation consistent with the volume rendering equations and may be harder to optimize; it may also makes it harder to separate volumetric ``occupancy'' from color.} The attenuation or accumulated transmittance along the ray corresponds to the optical path length, and is given simply by, 
\begin{equation}
T(t) = \exp\left(-\int_{t_n}^{t} \sigma(\bm r(s))\,ds\right).
\end{equation}
We refer to the original papers for details on how to discretize and solve these equations numerically to obtain the computed color $\hat{C}(\bm r)$.  The optimization simply minimizes the error between the rendered and true pixel colors:
\begin{equation}
\mathcal{L} = \sum_{\bm r \in R} \left\| \hat{C}(\bm r) - C(\bm r)  \right\|_2 ,
\end{equation}
where $R$ is the set of rays, and we simply compare ground truth (acquired images) $C(\bm r)$ and our reconstructed colors $\hat{C}(\bm r)$. 

\begin{figure}
\includegraphics[width=\columnwidth/2]{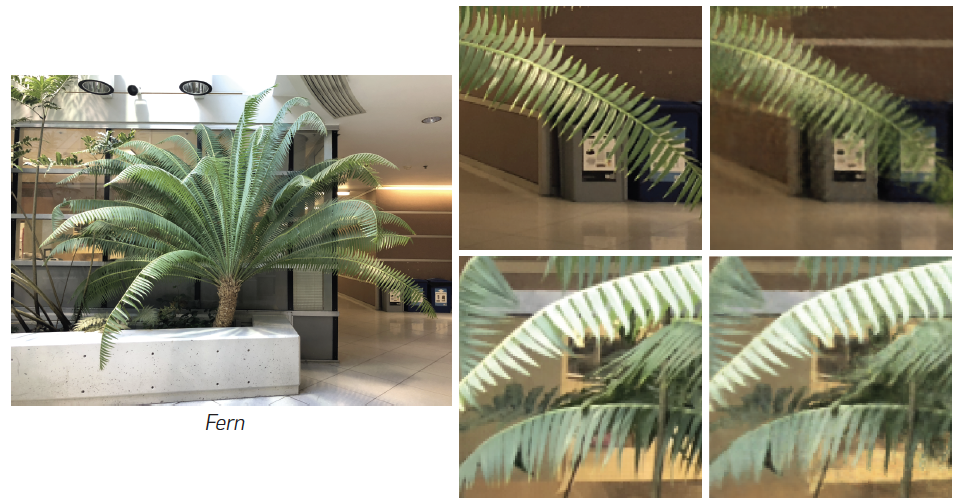}
\includegraphics[width=\columnwidth/2]{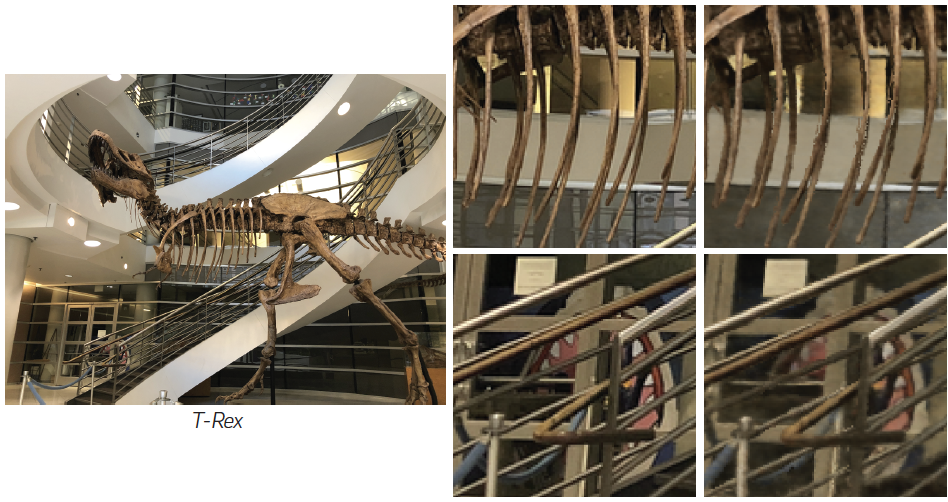}
\caption{\em Real results obtained by the original NeRF algorithm, showing a close resemblance to ground truth.  The ground truth is on the left of the insets, with the NeRF algorithm on the right.}
\label{fig:results}
\end{figure}

While this is a simple formulation, it is important to note the physical limitations.  The theory of volumetric scattering or radiative transfer~\cite{Chandrasekhar} is well known in astrophysics and computer graphics, among many other fields.  We have taken only a subset of it, omitting scattering entirely, but keeping absorption and emission.  Moreover, most real scenes are not actually emissive and in many cases not really volumes (rather surfaces).  However, the volumetric representation is critical, as we will discuss later.  Even though the scene is not emissive and actually obtained by an interplay of lighting/illumination and the reflectance properties of the objects, if we fix the lighting, we can ``bake in'' this illumination and effectively treat the scene as emitting light.  However, it is to be understood that these are mathematical conveniences rather than a correct physical model.  

Finally, the original papers introduced a positional encoding, which is crucial for an MLP to recover high-frequency functions and produce high-quality results, a topic which we studied in more depth in a companion work~\cite{Tancik}.
More specifically, the inputs are mapped to a higher-dimensional space, in essence Fourier functions at a variety of higher frequencies, making it easier for the network to learn these higher frequencies.  Consider a single input $p$ in the real numbers $\mathcal{R}$ (the same mapping can be applied separately to Cartesian components of spatial position and direction).  We lift this input to a higher-dimensional space $\mathcal{R}^{2L}$ with the encoding, 
\begin{equation}
\gamma(p) = \left(\sin(2^0 \pi p),\cos(2^0 \pi p), \ldots , \sin(2^{L-1}\pi p), \cos (2^{L-1}\pi p)\right).  
\end{equation}

Putting this all together, some results for view synthesis on real scenes for the original NeRF paper are shown in Fig.~\ref{fig:results}, where the quality closely matches the ground truth (input images).  

The original NeRF algorithm had many limitations, in terms of excessive training and rendering times, optimizations for one scene/object at a time, requirements of tens to a hundred views, limitations to static bounded scenes, only view synthesis without lighting changes or recovering material parameters, and many more.  At least initial work on addressing all of these limitations has proceeded with more than 3,000 papers from research groups around the globe having built on the original NeRF method to date as of July 2023, addressing alomst every conceivable application in graphics, vision and beyond.  NeRFs have been used in applications relating to robotics~\cite{Kerr}, tomography~\cite{Ruckert}, and astronomy~\cite{blackhole}. A number of websites\footnote{\url{https://dellaert.github.io/NeRF/} and \url{https://neuralradiancefields.io/} among many others.} have been written and maintained by independent authors, and there are independent blogs and articles summarizing and discussing the large number of NeRF papers at recent vision conferences.\footnote{ICCV 21: \url{https://dellaert.github.io/NeRF21/}, CVPR 22: \url{https://dellaert.github.io/NeRF22/}, NeurIPS 22: \url{https://markboss.me/post/nerf_at_neurips22/}, ECCV 22: \url{https://markboss.me/post/nerf_at_eccv22/}, CVPR 23: \url{https://markboss.me/post/nerf_at_cvpr23/}.  The reader who seeks to follow the recent NeRF literature can follow the links and excellent summaries on these pages.}  Almost every subsequent vision, learning, and graphics conference has given at least one of their best paper awards\footnote{A subset of the original NeRF authors are the drivers or some of this work~\cite{mipnerf,Ishiteccv,dreamfusion,refnerf} but the majority of these and related papers come from a diverse set of research groups around the world.} to articles based on NeRF and extensions, for example, Giraffe at CVPR 21~\cite{Giraffe}, MipNeRF and Coco3D at ICCV 21~\cite{mipnerf,Coco3D}, RefNeRF at CVPR 22~\cite{refnerf}, our own work on Neural Implicit Evolution at ECCV 22~\cite{Ishiteccv}, InstantNGP from SIGGRAPH 22~\cite{instantNGP}, Dynamic image-based rendering from CVPR 23~\cite{dynibar}, 3D Gaussian Splatting for Real-Time Radiance Field Rendering from SIGGRAPH 23~\cite{Kerbl}, and DreamFusion at ICLR 23~\cite{dreamfusion}.  This is likely not an exhaustive list.

The subsequent InstantNeRF method from NVIDIA~\cite{instantNGP} has addressed many shortcomings in terms of rendering and training time, often enabling training in under a minute and rendering at real-time frame rates, leading to recognition by Time Magazine as one of the best inventions of 2022.\footnote{\url{https://time.com/collection/best-inventions-2022/6225489/nvidia-instant-nerf/}}  This has enabled a number of applications, including creation of NeRFs by lay users; some artist creations are showcased in an Instant NeRF AI Gallery website.\footnote{\url{https://www.nvidia.com/en-us/research/ai-art-gallery/instant-nerf/}}  The New York Times has used NeRFs for capturing portraits,\footnote{\url{https://rd.nytimes.com/projects/creating-workflows-for-nerf-portraiture}} and Mark Zuckerberg, CEO of Meta/Facebook has mentioned NeRFs and Inverse Rendering as key technologies for the metaverse.\footnote{\url{https://youtu.be/hvfV-iGwYX8?t=4400}}  Given NeRF's popularity, a number of software frameworks have been developed; one open source example is NeRFStudio~\cite{nerfstudio}.  A number of surveys on NeRF have been published~\cite{Dellaertsurvey,Tewari2022STAR} which we refer readers to, and even the popular press has highlighted the method~\cite{wiredarticle}. Our goal is not to duplicate these writeups, and indeed it is impossible to do justice to all the excellent follow-on work in a few pages.

The remainder of this article provides historical context on the three-decades long quest for the ``right'' 3D scene representation for view synthesis or image-based rendering.  We then discuss new developments in core NeRF representations, and relate them to earlier work on light transport reprsentation and real-time rendering.  Our goal is not to be exhaustive, but identify some of the core challenges and solutions in the NeRF algorithm.  


\section{Scene Representations for Image-Based Rendering}

Our story begins at least 30 years ago, pre-dating the research careers (and for students indeed their birth) of any of the authors of the NeRF paper.  In SIGGRAPH 93, Chen and Williams~\cite{ChenWilliams} first introduced the notion of view interpolation for image synthesis.\footnote{Lance Williams received SIGGRAPH's 2001 Coons Lifetime Achievement Award, the community's highest honor, primarily for pioneering the image-based approach to graphics or ``brute force in image space'' including shadow mapping~\cite{Williams78} and mip-mapping~\cite{mipmapping}.}  This paper has the nearly unique distinction of being included in both volumes of seminal graphics papers~\cite{seminal2,seminal1} produced by SIGGRAPH for the 25$^{\text{th}}$ and 50$^{\text{th}}$ anniversary conferences.
While Chen and Williams used synthetic z-buffer images, this paper is widely taken as defining the view synthesis problem, and leading to the immensely popular field of image-based rendering (IBR) in computer graphics and vision in the 1990s.  

At that time, computer graphics rendering or image synthesis had already achieved striking realistic imagery but was limited by the quality of input models for geometry, lighting and materials.  Image-based rendering or view synthesis offered an alternative of using by definition photorealistic input images and combining them to create new images.  Within computer vision, this was an exciting new problem direction, as opposed to the traditional problems of 3D reconstruction or recognition.  Indeed, much of the early promise in image-based rendering was to bypass error-prone 3D reconstruction entirely.  Tomaso Poggio's group pioneered work on facial analysis and synthesis in an image-based way (for example~\cite{Poggio1}), but often does not get the credit richly deserved since these publications were not in the mainstream graphics community.

Much like the explosion of work on NeRFs, there was an explosion of work on view synthesis and image-based rendering, including classic papers on plenoptic modeling~\cite{McMillan}, light field rendering~\cite{Levoy}, the lumigraph~\cite{Gortler}, layered-depth images~\cite{Shade}, and thousands of others we cannot cover in this article.  Indeed, the seminal graphics papers volume from the 50$^{\text{th}}$ SIGGRAPH conference in 2023~\cite{seminal2} includes at least a dozen IBR papers from the mid-1990s to the mid-2000s (listed in chronological order)~\cite{ChenWilliams,McMillan,Levoy,Gortler,Deb1,environmentmatting,Debevec1,Wood,ulr,NayarGI,phototourism}.  I highly recommend re-reading these classics. The SIGGRAPH technical awards program and committee did a remarkable job keeping up with these advances, with major computer graphics technical awards deservedly given to at least one author of essentially all of the above-mentioned papers.  

Subsequent developments have especially been inspired by the light field\footnote{Some seminal early work on the plenoptic function or light field comes from Adelson~\cite{Adelson}, although the concept of light fields and light field cameras goes back over a century, pre-dating computing~\cite{Ives,Lippman}.} and lumigraph papers~\cite{Gortler,Levoy} which show how resampling and combining rays can achieve view synthesis, and more importantly provide the foundations for reconstructing the light field from sparse images.  Indeed, my own interest in the project that eventually led to NeRF started with depth reconstruction and view interpolation from light field cameras~\cite{Nima,TaoUnified}, using Lytro cameras from co-author Ren Ng's startup.  Layered-depth images or LDIs~\cite{Shade} are one precursor to Multi-Plane Images or MPIs widely used prior to the NeRF method (including in our work on local light field fusion~\cite{llff}).  

A critical question in these and follow-on works has been the right 3D scene representation for image-based rendering.  Some early papers like the original ight field work~\cite{Levoy} argued that we did not need a scene representation; one was simply interpolating images.  In this way, the error-prone 3D reconstruction step could completely be skipped.  However, in a seminal paper at SIGGRAPH 2000, Chai et al.~\cite{Chai} formally demonstrated a plenoptic sampling theory, showing that depth information can enable much sparser acquisition, and in fact there is a tradeoff between depth accuracy and sampling rate (an analysis we exploited in our own subsequent work~\cite{llff}).  The lumigraph paper~\cite{Gortler} had already explored constructing an approximate geometric representation.  

Ultimately, it became clear that despite the name ``image-based rendering'', the key element was the intermediate 3D scene representation.\footnote{Readers are encouraged to read the ACM Dissertation Award honorable mention theses of the NeRF authors; in particular Pratul Srinivasan's thesis~\cite{Pratulthesis} discusses his quest for the best representation with deep learning for view synthesis.}  In essence, most algorithms follow the pipeline of Fig.~\ref{fig:setup}, where one takes the input views, optimizes an intermediate 3D scene representation, and then renders the final views.  Inspired by traditional computer graphics and vision, it is natural for the scene representation to be approximate 3D geometry such as a rough mesh, or perhaps a disparity map.  But these representations can be very challenging for high-quality view synthesis.  For example, consider the canonical voxel-coloring approach by Seitz and Dyer~\cite{Seitz97}.  Once a mistake is made, the visibility relationships in the shapes are impacted going forward and the final recovered 3D structure is inaccurate with incorrect visibility.  Moreover, if geometry is incorrect, then view-dependent effects like specularities are often incorrect.  Nevertheless, the early image-based rendering work achieved a certain level of success with increasingly sophisticated algorithms over the next two decades; representative papers are~\cite{Chaurasia} that uses more conventional depth synthesis and warping, and~\cite{Penner} that acknowledges the limitations of depth and 3D reconstruction, aiming for a notion of a ``soft'' or ``probabilistic'' 3D reconstruction that does not commit to a hard 3D shape.  In effect, along a given ray, we are not considering a single surface or intersection points, but a probability distribution of intersections.  This is a very important insight, which NeRFs in some sense take to its natural conclusion.  

In 2012, Krizhevsky et al.~\cite{Krizhevsky} introduced massive gains in classification problems with deep convolutional networks, leading to a tsunami of work in deep learning for computer vision.  For a while, it seemed that image synthesis would not be touched by these developments.  However, fully convolutional networks and U-nets enabled creation of high-resolution imagery~\cite{Shelhammer} and results were quickly applied to stereo reconstruction~\cite{Flynn}.  Our own work considered view synthesis from light-field cameras, obtaining a fast algorithm to obtain the full light field from only four corner views~\cite{Nima}.  This paper (co-authored with my postdoc Nima Kalantari and Ph.D. student Ting-Chun Wang) can reasonably be considered the first effective deep-learning algorithm for view synthesis, and was a critical marker for the research that followed.  Nevertheless, the representation proposed still largely used disparities and warping in the classical sense.  

Our paper on Neural Radiance Fields presents a dramatic break from this tradition of 3D geometric representations.  First, the representation of the scene is a volume rather than a surface.  Second, it is a continuous volume rather than a discrete voxel grid.  It is likely that this core insight of a continuous volumetric representation is the crucial lasting insight from the NeRF paper, although there are also a number of contributions in the specific learning algorithms.  Of course, research doesn't happen in a vacuum, and there were a number of important precursors.  Volume rendering~\cite{volumerendering} has a long history in graphics, and complex materials are increasingly represented in a volumetric fashion~\cite{microflake}.  Indeed, I had a large NSF grant starting in 2010 to propose a fundamentally new volume-based material representation; the NeRF paper a decade later has truly made volumes a first-class primitive for scene representation.  We were inspired by the neural volumes work~\cite{neuralvolumes} that uses a discrete voxel grid with a few other limitations, predicted using a deep 3D CNN (a similar idea is proposed in Deep Voxels~\cite{Sitzmann}), but NeRF generalizes this idea in many ways and in particular to a {\em continuous} rather than discrete volumetric representation.  As an aside, instead of simply storing volume densities, one can store reflectance to recover light and view-dependent properties, a line of work my group pursued concurrently~\cite{Sainrf,refvol}.  

In terms of learning algorithms also, our NeRF paper breaks new ground.  Instead of using a CNN, the dominant paradigm at the time, one optimizes a simple MLP; the original work required only 5MB for the weights of this fully-connected network to represent the entire scene.  Again, the idea is not entirely new, and builds on pioneering work on deep signed distance functions~\cite{DeepSDF} and scene representation networks~\cite{Sitzmann2}.  Finally, NeRF introduces the notion of positional encoding for high-frequency features, analyzed in more detail in a companion paper~\cite{Tancik}, and built upon in a wide range of applications by many authors.  This leverages work in the learning community on spectral bias in neural networks~\cite{Rahaman}.  It should be noted, as discussed in the next section, that a number of new NeRF representations have modified the learning approach to use only shallow MLPs without positional encoding, and in the most recent work~\cite{Kerbl} not even using a neural representation.  However, the core concept of a continuous volumetric representation, introduced in NeRF, has pervaded essentially all of the subsequent work.  

\section{Representations for Neural Radiance Fields}

The original NeRF paper has spawned an explosion of new representations, which we briefly discuss here (but this is far from an exhaustive list).  Though rarely cited, this parallels the development two decades ago of compact representations for light transport and precomputed radiance transfer (PRT)~\cite{Sloan1}, in which the author of this article played a major role~\cite{prtsurvey}.  For example, spherical harmonics and reflection-based reparameterizations are commonly used~\cite{refnerf}.  An early method in the PRT literature was to cluster scenes~\cite{Sloan3} or to break images into blocks~\cite{Nayar1}.  Similarly, KiloNeRF~\cite{Reiser} breaks the scene into regions, each fit with small MLPs.  However, there has so far been little analysis or understanding of optimal block and cluster sizes, or the tradeoffs involved, work which has been pursued in some depth in the PRT literature~\cite{Dhruv}.  One strong recommendation we make for future authors is to be aware of the PRT literature, and cite it/build on it appropriately.  There is also potential in the other direction, of bringing NeRF representations into classical graphics problems; we have made a first step towards this using NeRF-like ideas for PRT with glossy global illumination~\cite{prtraghavan}.

Many recent methods are based on the notion of feature-fields.  This makes a step back towards discrete volumetric representations, wherein a coarse volumetric grid of abstract features is stored explicitly.  One can then interpolate this grid at any point along the ray traversal to obtain features, which are subsequently fed into a small MLP to decode the color and density values, after which rendering proceeds like in the original NeRF algorithm.  While this does take considerably more memory than a standard NeRF (but much less than a full-resolution voxelized volumetric model), it is easier to train and substantially faster to evaluate, given the smaller MLPs required.  One challenge then is the appropriate representation and compression of the feature field itself.  Two recent representative works in this area are Instant Neural Graphics Primitives~\cite{instantNGP} which uses a multiresolution hash-grid and TensorRF~\cite{Chen2022ECCV} which does a tensor deocmposition of the feature grid.  Note that TensorRF has strong similarities to clustered tensor approximation~\cite{Tsai} in PRT, and we encourage future authors to more carefully investigate the potential for clustering and analyze the number of terms needed in this representation.  It is also possible to combine TensorRF for factorization and InstantNGP for representation of the lower-dimensional components with multiresolution hash-grids; this idea has been used for example for dynamic full-body human models~\cite{HumanRF} and for static PRT in our own work~\cite{prtraghavan}.    In any event, this explosion of new representations, factorizations and compressions indicates that mathematical representations and signal-processing are key even in the deep learning-NeRF era.  

There have also been a few other novel representations proposed.  EG3D~\cite{eg3d} essentially projects a 3D point to lower-dimensional 2D planes.  Since the 2D planes are lower-dimensional, they can represent much higher-resolution information explicitly as a simple texture-map, and CNN or transformer-based algorithms can even predict these representations directly~\cite{realtimeradiance}.  This can be extended to higher dimensions using K-planes~\cite{kplanes}, which can even operate without any neural layers or MLP (earlier work on plenoxels developed an adaptive octree voxel representation, again without neural components~\cite{plenoxels}).  

The breakthrough in NeRFs was achieved by leveraging older computer graphics notions of volume rendering.  In the past year, further breakthroughs have been achieved through an even older graphics technique of point-based rendering, where points are used as a display primitive~\cite{levoywhitted,qsplat}.  Point-nerf~\cite{xu2022point} introduced the notion of point-based NeRF representations as easier to train.  A radical new approach was proposed by Kerbl et al.~\cite{Kerbl} using 3D gaussian splats for their point primitives.  This is currently the state-of-the-art method in terms of visual quality, rendering time and performance, but does have higher memory requirements.  In a sense, this work comes full-circle in the memory-speed tradeoff with a fully explicit volumetric representation.  Most remarkably, the technique is entirely classical, with no neural primitives, and the continuous volume essentially represented adaptively with gaussian blobs or splats, which serve as the point samples.  Within the past decade, this is perhaps the first time a major neural algorithm has been superseded by a non-neural classical equivalent.  
Again, research doesn't happen in a vacuum, and a similar Gaussian splat representation for surfaces and volumes in differentiable rendering for somewhat different shape from silhoutte and related problems was proposed a year earlier by Keselman and Hebert~\cite{Keselman}.

Which brings us back to a key point made in the last section: While NeRFs are often thought about as another major victory for AI-based techniques in making remarkable advances on decades-old problems, the key insight may actually simply be in the idea of a continuous volumetric representation,\footnote{It can be argued that the point-based representation in Gaussian splats~\cite{Kerbl} is not strictly a volume and does not actually require ray marching, but the alpha compositing of splats obeys the same equations and we view it an adaptive representation of the continuous volume.}  instead of surfaces or discrete voxel grids, rather than any particular learning representation or algorithm.  In fact, there may be no need for any neural component at all, and what we really care about is just the radiance field.

We have devoted a few paragraphs to a subject that has involved hundreds or thousands of papers over the past three years.  The field of NeRF representations is rapidly evolving, with many chapters still to be written in the years to come.  As such, NeRFs can be seen as one milestone, but by no means the final word on the ``best'' 3D scene representation for image-based rendering.  

\section{Conclusion: NeRFs as the new Geometric Primitive}
A representation of 3D geometry is critical within computer graphics.  Indeed, the canonical graphics pipeline involves 3D scene geometric models, along with lights, materials, animations and a rendering algorithm to create images.  Geometry is also manipulated by smoothing it, simulating it, meshing it, or using it as a representation for perception, recognition or editing.  To date, the most successful representation of 3D geometry in computer graphics is perhaps simply the triangle mesh.  But the history of computer graphics has had many rich alternatives, including spline patches, subdivision surfaces, implicit surfaces, and points.  

In conclusion, we argue that NeRFs are the new geometric primitive, based on volumes rather than surfaces, and based on neural rather than explicit representations (although we have seen some work in the last paragraph can achieve state-of-the-art performance without any neural primitives).  As such, NeRFs and neural implicit functions can often be a plug-in replacement for any problem that requires a 3D geometric representation.  This has led to their immense popularity, since the ideas can be applied to almost any domain.  It remains an open question as to whether operations in the computer graphics pipeline like physical simulation and geometry processing that have historically been applied on finite element models, meshes and tetrahedra should be extended to the NeRF domain, or whether it is simpler to extract an explicit mesh representation from a NeRF.  Regardless, the view of NeRFs as the new 3D geometric primitive is a valuable one, and enables their potential usage anywhere one seeks a 3D (or higher-dimensional) geometric model.  This gives us an immense set of applications to focus on, and we expect research on both core radiance field representations and NeRF applications to continue for many years to come.  

\section*{Acknowledgements: } 
First, I of course want to thank my co-authors on the NeRF papers, my Ph.D. student Pratul Srinivasan, Ben Mildenhall, Matt Tancik, Jon Barron, and Ren Ng.  We also acknowledge Kevin Cao, Guowei Yang and Nithin Raghavan for comments and discussions.  I wish to acknowledge as well the early students in the light field projects, including Michael Tao, Ting-Chun Wang and my postdoc Nima Kalantari, who first showcased the use of deep-learning based view synthesis for light fields.  I am grateful to all of my colleagues and students within the UC San Diego Center for Visual Computing.  I wish to thank the many funders of our research over the years; the original papers acknowledge ONR grants N000141712687, N000141912293, N000142012529 and most recently N00014-23-1-2526 and NSF Chase-CI grants 2100237 and 2120019, the Ronald L. Graham Chair, as well as NSF (Pratul Srinivasan, Matt Tancik) and Hertz Fellowships (Ben Mildenhall), compute credits through the BAIR commons program, and Blend Swap users for models.  I also acknowledge many gifts and industrial contracts over the years to our light field research program from Google, Adobe, Samsung, Qualcomm, Sony, and many other partners.  In particular, I would like to acknowledge program manager Behzad Kamgar-Parsi at ONR for a now close to two-decades long support of my research endeavors in physics-based computer vision, including the light field and NeRF projects.  

It is with a deep gratitude that we thank the organizers of the International Congress on Basic Science for recognizing Neural Radiance Fields with an inaugural Frontiers of Science Award, and for their support of basic scientific research in so many fields.  NeRFs represent the culmination, or perhaps a waypoint and milestone, in a three-decades long quest to find the best scene representation for view synthesis and image-based rendering.  Perhaps the greatest thanks go to the many researchers who initiated this research effort and sustained it over the past 30 years, and the many thousands of authors who have built on the original NeRF paper in the past three years.  The Frontiers of Science Award really belongs to them all.  

\bibliography{combined}
\bibliographystyle{plain}

\end{document}